\documentclass{article}
\usepackage{ijcai22}

\usepackage{times}

\usepackage{soul}
\usepackage{url}
\usepackage[hidelinks]{hyperref}
\usepackage[utf8]{inputenc}
\usepackage[small]{caption}
\usepackage{graphicx}
\usepackage{amsmath}
\usepackage{booktabs}
\usepackage{enumitem}
\usepackage{balance}

\usepackage{amsfonts}
\usepackage{multirow}
\usepackage{booktabs}
\usepackage{colortbl}
\usepackage{algorithm}
\usepackage{enumitem}
\usepackage{appendix}
\usepackage{algpseudocode}
\usepackage{caption}
\usepackage{bm}
\usepackage{newunicodechar}
\usepackage[table]{xcolor}

\usepackage{caption}
\newunicodechar{ﬁ}{fi}
\newunicodechar{ﬀ}{ff}
\newunicodechar{ﬂ}{fl}
\newcommand{\ie}{\emph{i.e., }}
\newcommand{\eg}{\emph{e.g., }}

\newcommand{\wrt}{\emph{w.r.t. }}
\newcommand{\cf}{\emph{cf. }}

\title{Graph Structure Learning with Bi-level Optimization}

\author{Nan Yin
}

\begin{document}

\maketitle

\begin{abstract}

Currently, most Graph Structure Learning (GSL) methods, as a means of learning graph structure, improve the robustness of GNN merely from a local view by considering the local information related to each edge and indiscriminately applying the mechanism across edges, which may suffer from the local structure heterogeneity of the graph (\ie the uneven distribution of inter-class connections over nodes). To overcome the cons, we extract the graph structure as a learnable parameter and jointly learn the structure and common parameters of GNN from the global view. Excitingly, the common parameters contain the global information for nodes features mapping, which is also crucial for structure optimization (\ie optimizing the structure relies on global mapping information). Mathematically, we apply a generic structure extractor to abstract the graph structure and transform GNNs in the form of learning structure and common parameters. Then, 
we model the learning process as a novel bi-level optimization, \ie \textit{Generic Structure Extraction with Bi-level Optimization for Graph Structure Learning (GSEBO)}, which optimizes GNN parameters in the upper level to obtain the global mapping information and graph structure is optimized in the lower level with the global information learned from the upper level. We instantiate the proposed GSEBO on classical GNNs and compare it with the state-of-the-art GSL methods. Extensive experiments validate the effectiveness of the proposed GSEBO on four real-world datasets.
\end{abstract}

\section{Introduction}
\label{introduction}

Based on the \textit{homophily assumption} of ``like to associate with like" ~\cite{mcpherson2001birds,yin2023messages,shou2023adversarial},
Graph Neural Network (GNN) \cite{scarselli2008:graph,yin2024continuous,yin2024dynamic,ju2024survey} 
has become the promising solution for node classification. 
However, a large portion of edges are inter-class connections~\cite{netprobe}, and representation propagation over such connections can largely hinder the GNN from obtaining class-separated node representations, hurting the performance.

\begin{figure}
  \centering
  \setlength{\abovecaptionskip}{0cm}
  \setlength{\belowcaptionskip}{0cm}
  \includegraphics[scale=0.72]{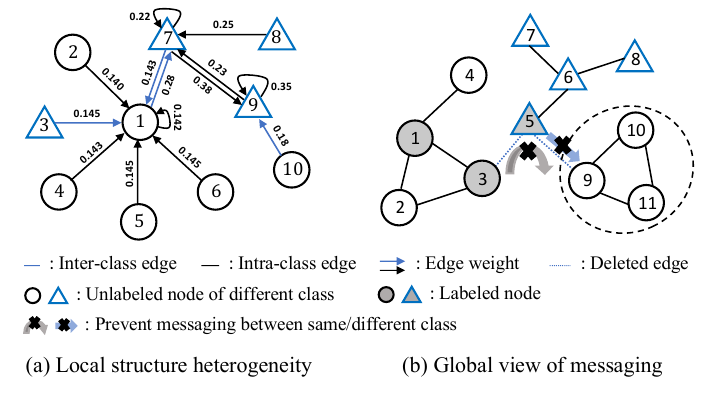}
  \caption{Examples for (a) the impact of local structure heterogeneity (\textit{node 1} and \textit{9}, the weights between nodes do not provide enough information for classes differentiation), and (b) the global view of messaging (the edge between \textit{node 3} and \textit{node 5}, the deleted edge prevents the intra-class information transfer).
  }
  \vspace{-0.2cm}
  \label{attention}
\end{figure}

Existing GSL methods are roughly categorized into attentive mechanism, noise detection, and probabilistic mechanism.
  \textit{\textbf{Attentive mechanism}}, which calculates weights for edges to adjust the contribution of different neighbors during representation propagation~\cite{velivckovic2017graph,wang2021multihop,JiangZLTL19,WZ20,ZhaoWSHSY21}. These methods can hardly work well in practice for two reasons: 
  (1) the mechanism may not generalize well to all nodes with different local structures (\cf Figure \ref{attention}(a)); 
  and (2) the attention cannot be easily trained well due to the limited labeled data~\cite{knyazev2019understanding}.
   \textit{\textbf{Noise detection}}, which incorporates an edge classifier to estimate the probability of inter-class 
  connection for each edge~\cite{zheng2020robust,zhao2020data,luo2021learning,kazi2020differentiable,franceschi2019learning,LiWZH18}. 
  Although it can be better trained owing to the supervision of edge labels, 
  the edge classifier also suffers from local structure heterogeneity and lacks consideration of global information.
  \textit{\textbf{Probabilistic mechanism}}, which models connection from a global view which assumes a prior 
  distribution of edge and estimates GNN parameters with Bayesian optimization to overcome the impact of inter-class edges 
  \cite{zhang2019bayesian,elinas2019variational,wang2020learning,WuRLL20}.
  Although the edge specific parameterization largely enhances the model representation ability, 
  it is hard to accurately access the prior distribution.
  
Despite the achievements of the existing methods, there still exists some common cons: (1) \textbf{Edge modeling method}, the existing methods model edges with parameter sharing mechanism, which may suffer from the local structure heterogeneity problem; (2) \textbf{Local optimization}, the local optimization problem focuses on optimizing the parameters with the information of neighbor nodes, which ignores the impact from the global view.  
Therefore, we come up with the key considerations for GSL: 
(1) modeling graph connection in an edge-specific manner instead of a shared mechanism; and (2) optimizing the corresponding parameters with a global objective of accurately classifying all nodes. 
The edge-specific modeling can overcome the local structure heterogeneity, 
\ie handling nodes with different properties (\eg \textit{node 1} and \textit{node 9} in Figure~\ref{attention}(a)) via different strategies.
Besides, blindly removing the inter-class edges will increase the risk of misclassifying the target nodes (in dash circle) due to cutting off their connections to the labeled neighbors in the same class (\eg edge between \textit{node 5} and \textit{node 9} in Figure~\ref{attention}(b)). Thus, it is necessary to optimize the graph structure from the global view instead the local ones.

However, it is non-trivial to achieve the targets mentioned above due to the following challenges: 
(1) the graph structure is embedded into the GNN model, which affects the procedure of model parameter optimization once updated, requiring a careful design of the model training algorithm;
(2) the calculation of the ideal global objective is intractable due to the limited labelled nodes, especially 
the hard semi-supervised setting. 

In this work, we propose a new \textit{Generic Structure Extraction with Bi-level Optimization for Graph Structure Learning (GSEBO)}, which optimize the graph structure and learn the node embeddings from the global view. 
In particular, we first devise a new generic structure extractor, which accounts for the graph structure with both the connectedness between nodes and the strength of connections. In addition to the adjacency matrix, the extractor adopts a learnable matrix to represent the graph structure and adjusts the representation propagation. Moreover, we design a bi-level optimization algorithm where the outer and inner optimizations update the structure and the parameters of the base graph convolution (\eg feature mapping parameters). In this way, we decompose the hard optimization issue of GSEBO into two easy ones. In addition, we set the objective of outer optimization as the validation loss to better approximate the ideal global objective.
The proposed generic structure extractor can be extended to most existing graph convolution operators. We instantiate it on four representative GNN models (\ie GCN~\cite{kipf2016semi}, GAT~\cite{velivckovic2017graph}, GraphSAGE~\cite{hamilton2017inductive}, and JK-Net~\cite{xu2018representation}) and compare GSEBO with state-of-the-art GSL methods, which are evaluated on four real-world node classification datasets. Extensive experiments justify the rationality, effectiveness and robustness of the proposed method. 
In summary, our main contributions are as follows:

\begin{itemize}
  \item We propose a novel \textit{GSEBO} with bi-level optimization for edge-specific graph structure learning, which learns the graph structure from a global view by optimizing a global objective of node classification.
  \item We devise a generic structure extractor, which parameterizes the strength of each edge during representation propagation. Besides, we summarize how the classical GNN methods are transferred in the form of learnable graph structure with generic structure extractor.  
  \item We evaluate the proposed GSEBO with classical GNNs as backbones and compare it with the state-of-the-art GSL methods. Extensive experiments on four real-world datasets show the superior learning ability of the proposed method over the existing methods.
\end{itemize}

\section{Related Work}

\begin{figure*}
  \centering
  \setlength{\abovecaptionskip}{0cm}
  \setlength{\belowcaptionskip}{0cm}
  \includegraphics[scale=0.85]{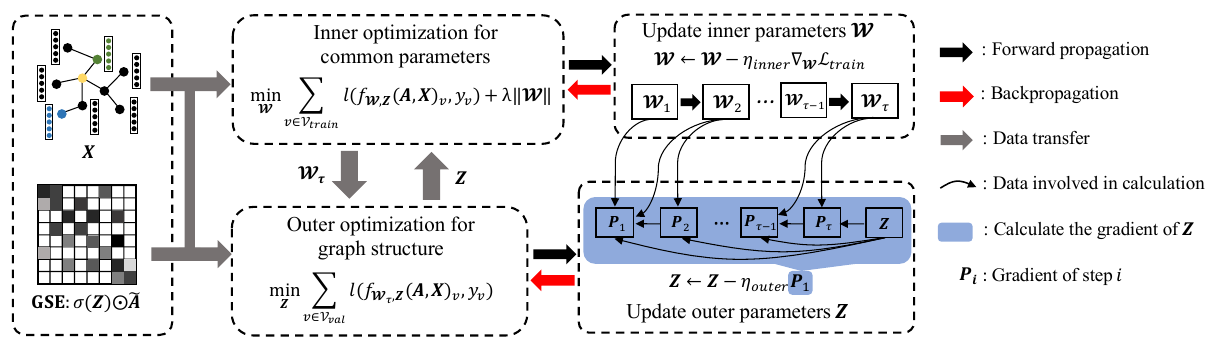}
  \caption{The framework of GSEBO. The generic structure extractor (GSE) decouples the graph structure from GNNs to account for the edge-specific modeling, the inner and outer optimization steps are adopted for common parameters and graph structure optimization.}
  \vspace{-0.1cm}
  \label{fig1}
\end{figure*}


\paragraph{Attentive mechanism.}
The attentive mechanism methods adaptively learn the weights of edges and adjust the contributions of neighbor nodes~\cite{pang2023sa,yin2022dynamic,yin2023dream}. 
MAGNA \cite{wang2021multihop}  incorporates multi-hop context information into every layer of attention computation. 
IDGL \cite{WZ20} uses the multi-head self-attention mechanism to reconstruct the graph, which has the ability to add new nodes without retraining.
HGSL \cite{ZhaoWSHSY21} extends the graph structure learning to heterogeneous graphs, which constructs different feature propagation graphs and fuses these graphs together in an attentive manner.
However, those methods suffer from different local structures and are difficult to train.

\paragraph{Noise detection.}
The noise detection methods leverage the off-shelf pre-trained model to induce node embeddings or labels and incorporate an edge classifier to estimate the probability of each edge~\cite{yin2023omg,yinsport}. 
NeuralSparse \cite{zheng2020robust} considers the graph sparsification task by removing irrelevant edges. 
GAUG \cite{zhao2020data} utilizes a GNN to parameterize the categorical distribution instead of MLP in NerualSparse. 
PTDNet \cite{luo2021learning} prunes task-irrelevant edges by penalizing the number of edges in the sparsified graph with parameterized networks. 
Even though the noise detection methods can be well trained with the supervision of edge labels, the edge classifier also suffers from local structure heterogeneity and lacks consideration of global information.

\paragraph{Probabilistic mechanism.}
This type of methods assume the prior distribution of graph or noise and estimate the parameters through observed values, then resample the edges or noise to obtain a new graph~\cite{yin2023coco,zhang2019bayesian,yin2022deal}.
BGCN \cite{zhang2019bayesian} estimates the parameter distribution of edges and communities by sampling edges from graph, and resample new graphs with the estimated parameters for prediction. 
VGCN \cite{elinas2019variational} trains a graph distribution parameter similar to the original structure through ELBO, and resample graphs for prediction. However, both of BGCN and VGCN models are sampled from noisy graph, the estimated parameters also contain noise. 
DenNE \cite{wang2020learning} assumes the observed graph is composed of real values and noise and the prior distribution of features and noise is known. With a generative model, the likelihood is used to estimate the representation of nodes. 
However, this method highly relies on the priors of feature and noise, which is difficult to obtain accurately. 

\paragraph{Bi-level optimization on GNN.}
LDS \cite{franceschi2019learning} jointly learns the graph structure and GNN parameters by solving a bilevel optimization issue that learns a discrete probability distribution for each edges. According to the learned distributions, LDS generates a new graph structure by sampling. Towards this end, LDS sets the objective of outer optimization as generating the observed edges, which clearly has a gap to the overall classification objective.
Moreover, LDS needs to estimate $N^2$ distribution parameters at least, which is hard due to insufficient labels ($|\mathcal{E}| \ll N^2|$. 
On the contrary, our method only activates a small portion of entries in $\bm{Z}$ where $A_{ij} = 1$, \ie the number of estimated parameters is same as the number of edges in graph $\bm{G}$ (\ie $|\mathcal{E}|$).
\section{Methodology}
\label{UDF}

We first introduce the essential preliminaries for GNN, and then elaborate the \textit{graph convolution operator} and 
\textit{bilevel optimization algorithm} of the proposed GSEBO.

\subsection{Preliminary}
\label{Preliminary}

Let $\bm{G} = (\mathcal{V}, \mathcal{E}, \bm{X})$ represents a graph with $N$ nodes and $M$ edges,
where $\mathcal{V}=\{v_1,v_2,\cdots,v_N\}$ and $\mathcal{E}=\{e_1,e_2,\cdots,e_M\}$ denote the set of nodes and edges 
respectively. $\bm{X}=[\bm{x}_1,\bm{x}_2,\cdots,\bm{x}_N]^\top \in \mathbb{R}^{N \times C}$ are nodes features, 
where $\bm{x}_i \in \mathbb{R}^{C}$ is the $i$-th row of $\bm{X}$, corresponds to node $v_i$ in the form of 
a $C$-dimensional vector. The adjacency matrix $\bm{A} \in \{0,1\}^{N \times N}$ indicates the connectedness of node pairs.

\paragraph{Node classification.} This task aims to learn a classifier $f(\bm{A}, \bm{X};\bm{\theta})$ from a set of 
labeled nodes to forecast the remaining nodes labels, where $\bm{\theta}$ denotes model parameters. 
Assuming there are $M$ labels, we index them from 1 to $M$ without loss of generality. 
Formally, $\bm{Y}=[y_1,y_2,\cdots,y_N]^\top \in \mathbb{R}^{N}$ are the labels of nodes, 
where $y_i\in \mathbb{R}$ is the label of node $i$. The target is achieved by optimizing the model parameter $\bm{\theta}$ over the labeled nodes, which is formulated as:
\begin{align}\label{eq:train_obj}
    \min_{\bm{\theta}} \sum_{i \leq M} l(f(\bm{A}, \bm{X})_i, y_i;\bm{\theta}) + \lambda \|\bm{\theta}\|,
\end{align}
where $l(\cdot)$ is a classification loss, and $\lambda$ is a hyperparameter to adjust the strength of parameter regularization.



\subsection{Generic Structure Extraction with Bi-level Optimization}
To optimize the graph structure, the key consideration lies in (1) decoupling the graph structure from the GNNs to account for the edge-specific modeling and (2) learning the graph structure from the global information in $\bm{\theta}$. 

\paragraph{Generic structure extractor.} Towards the first purpose, the core idea is to decompose the graph structure information into connectedness (the edges in the adjacency matrix) and the strength of connection (the latent variable). 
In general, there are two ways to model the connection strength regarding whether relying on the inductive bias 
of translation invariant or not. On one hand, attentive mechanisms or noise detection models are translation invariant, which decode the connection strength of each edge from its local features. However, with the consideration of the local structure heterogeneity issue in most real-world graphs~\cite{xu2018representation}, it is risky to rely on the translation invariant bias.
On the other hand, probabilistic mechanisms separately model the connection strength for each edge, 
where each edge corresponds to a specific distribution. However, it is non-trivial to set a proper prior in practice. 
According to these pros and cons, we summarize two considerations for extending the graph convolution: 
(1) edge-specific modeling; and (2) optimization without prior.


Towards this end, we model the connection strength as a parameter matrix $\bm{Z}$ with the same size as $\bm{A}$. Formally, the generic structure extractor (GSE) is abstracted as:
\begin{align}
\textbf{GSE}(\bm{Z}):&=\sigma(\bm{Z}) \odot \tilde{\bm{A}},\notag\\
    \bm{H}^{(k)} = \text{COM}(\bm{H}^{(k - 1)}&, \textbf{AGG}(\bm{H}^{(k - 1)}, \textbf{GSE}(\bm{Z})),\notag
\end{align}
where $\sigma(\cdot)$ is a non-negative activation function since the value of strength is always positive\footnote{In this work, we use the \textit{min[{max{[0,x]},1}]} function to restrict the value within [0, 1].}. \textbf{COM} and \textbf{AGG} are the combination and aggregation functions respectively.

Noteworthy, different from GNNs, GSE decouples the graph structure from GNNs and treats it as a learnable objective. Besides, GSE is a generic extractor, which can be instantiated over most existing graph convolutions (\cf Appendix A). 
In this way, as long as learning the connection strength is set properly, GSEBO can downweight the neighbors with inter-class connection during representation propagation, reducing the impact of inter-class with bi-level optimization.

\begin{figure}
  \centering
  \setlength{\abovecaptionskip}{0cm}
  \setlength{\belowcaptionskip}{0cm}
  \includegraphics[scale=0.5]{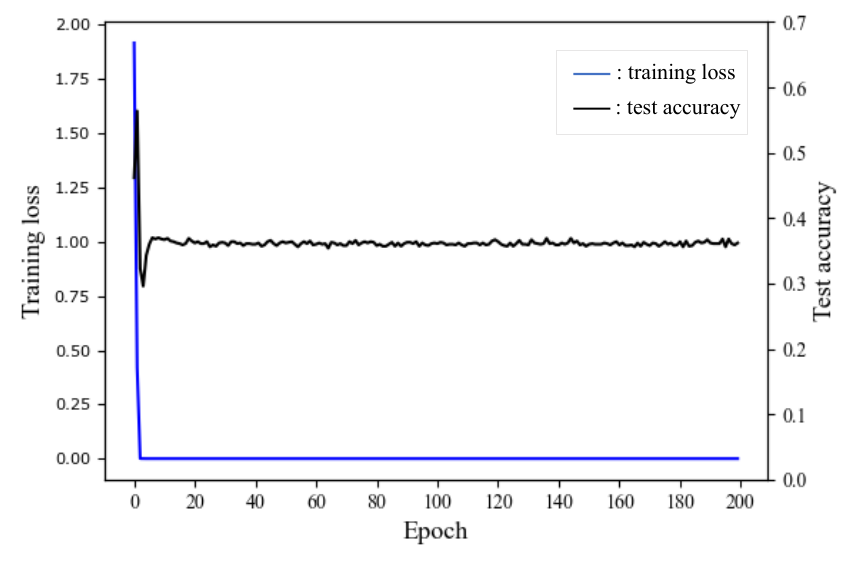}
  \caption{Optimize inner and outer steps on the training set of Cora.}
  \label{overfit}
\end{figure}

\paragraph{Bi-level optimization.} To achieve the second purpose of learning graph structure from global information, it is essential to carefully design a proper training algorithm to optimize the connection strength matrix $\bm{Z}$. Assume that we construct GSEBO with $K$ layers, which is denoted as $f(\bm{A}, \bm{X};\bm{\theta})$ with parameters $\bm{\theta} = \{\bm{Z}; \bm{W}^{(k)} | k \in [1, K]\}$. We have three main considerations for designing the training algorithm:
(1) Connection strength is a relative value, which changes across different views. As shown in Figure~\ref{attention}(b), the connection between \textit{node 3} and \textit{node 5} is weak from the local view, \ie the edge is inter-class and should be assigned a low weight. However, this edge is essential for the classification of \textit{node 9, 10, 11}, which deserves a high weight. Therefore, the optimization objective of $\bm{Z}$ should be the overall performance of node classification.
(2) $\bm{Z}$ and $\bm{\mathcal{W}} = \{\bm{W}^{(k)} | k \in [1, K]\}$ play different roles, but are closely related. The role of $\bm{Z}$ is close to $\bm{A}$, which restricts the model space for the mapping from node feature to label, and the role of $\bm{\mathcal{W}}$ includes the global mapping information for classification, which would relieve the cons of local optimization. That is, an update on $\bm{Z}$ will adjust $\bm{\mathcal{W}}$ and its optimization procedure. Therefore, the optimization of $\bm{Z}$ and $\bm{\mathcal{W}}$ are at two different but dependent levels.
(3) Directly minimizing the objective function of Eq. ~(\ref{eq:train_obj}) to obtain the parameters $\bm{Z}$ and $\bm{\mathcal{W}}$ is not able to achieve the desired purpose of learning the structure for the reason of over-fitting,
which is shown on Figure \ref{overfit}.

Towards this end, we propose a bilevel optimization \cite{domke2012generic,maclaurin2015gradient,franceschi2017forward} algorithm with inner and outer optimizations to learn $\bm{\mathcal{W}}$ and $\bm{Z}$, respectively. Figure~\ref{fig1} shows the overall procedure of the algorithm, where the inner and outer optimization steps are iteratively executed until stop condition.  
\begin{align}
   \min_{\bm{Z}}F(\bm{\mathcal{W}}^*(\bm{Z}))&\!=\!\sum_v l_{outer}(f(\bm{A},\bm{X})_v,y_v;\bm{\mathcal{W}}^*(\bm{Z})),\label{outer}\\
   s.t. \; \bm{\mathcal{W}}^*(\bm{Z})&\!=\!\arg\min_{\bm{\mathcal{W}}}L(\bm{\mathcal{W}},\bm{Z})\nonumber\\
   &\!=\!\sum_u l_{inner}(f(\bm{A},\bm{X})_u,y_u;\bm{\mathcal{W}},\bm{Z}).\label{inner}
\end{align}

 
 \begin{algorithm}[htb]
  \caption{Training of GSEBO.}
  \label{alg}
  \begin{algorithmic}[1]
    \Require
      $\bm{A}$: adjacency matrix;
      $\bm{X}$: nodes features;
      $\eta_{o},\eta_{i}$: outer and inner learning rates;
      $\tau$: number of inner steps;
    \State Initialize $\bm{\mathcal{W}}$ and $\bm{Z}$\footnotemark;

    \While{not \textit{early stopping}}
    \For{$t=1$ to $\tau$} \; \qquad \qquad \; \; \;  \# inner optimization;
    \State Update $\bm{\mathcal{W}}$ \wrt Eq. (\ref{eq:train_obj});
    \EndFor

    \State $\boldsymbol{\alpha}_{\tau}=\nabla_{\bm{Z}} \sum_{v \in \mathcal{V}_{val}} l(f(\bm{A}, \bm{X})_v, y_v;\bm{\mathcal{W}}_\tau)$;
    \State $\bm{P}=\bm{0}$;   \qquad \qquad  \; \; \; \; \; \qquad \,  \# initial outer gradient;
    \For{$\! t=\tau-1\!$ downto 1}   \; \# calculate outer gradient;
      \State $\bm{M}_{ t+1}=\frac{\partial \bm{\mathcal{W}}_{ t+1}}{\partial \bm{\mathcal{W}}_{ t}}$, $\bm{N}_{ t+1}=\frac{\partial \bm{\mathcal{W}}_{ t+1}}{\partial \bm{Z}}$;
      \State $\bm{P}=\bm{P}+\boldsymbol{\alpha}_{ t+1}\bm{N}_{ t+1}$;
      \State $\boldsymbol{\alpha}_{ t}=\boldsymbol{\alpha}_{ t+1}\bm{M}_{ t+1}$;
    \EndFor
    \State Update $\bm{Z}=\bm{Z}-\eta_{o}\bm{P}$; \qquad \; \; \;\# outer optimization;
   
    \EndWhile
    \setlength\belowdisplayshortskip{-10pt}
  \end{algorithmic}
\end{algorithm}

\footnotetext{We initialize $\bm{Z}$ with the normalized $\tilde{\bm{A}}$.}

\textbf{Inner optimization for common parameters.} 
This step is similar to the normal training of GNN for $\bm{\mathcal{W}}$ optimization. 
In particular, we update $\bm{\mathcal{W}}$ with a gradient descent based optimizer (\eg Adam~\cite{kingma2017adam}) over the training nodes by minimizing Eq.~(\ref{inner}) (\ie set $l_{inner}=l(f(\bm{A},\bm{X}),y;\bm{\theta})+\lambda \|\bm{\theta}\|$). 
When calculating the gradient of $\bm{\mathcal{W}}$, we treat the connection strength matrix $\bm{Z}$ as constant. 

Assuming that $\tau$ times of gradient descent, 
an approximate solution $\bm{\mathcal{W}}_1,\cdots,\bm{\mathcal{W}}_{\tau}$ of the inner optimization problem are obtained. 
For $t=1$ to $\tau$, the updated $\bm{\mathcal{W}}$ is calculated as  $\bm{\mathcal{W}}_{t}=\bm{\mathcal{W}}_{t-1}-\eta_{i} \nabla{L}(\bm{\mathcal{W}}_{t-1},\bm{Z})$, 
where $\eta_{i}$ is the inner learning rate, and the process of updating $\bm{\mathcal{W}}$ is shown on line 3-5 in $\textbf{Algorithm}$ 1.

\textbf{Outer optimization for graph structure.} 
Similarly, we treat the sequence of $\bm{\mathcal{W}}$ as constant to optimize the graph structure $\bm{Z}$. Ideally, the objective should be the overall classification loss of all nodes in graph. Formally,
\begin{align}
    \min_{\bm{Z}} \sum_{i \leq N} l(f(\bm{A}, \bm{X})_i, y_i;\bm{\theta}).\notag
\end{align}

Apparently, the calculation of the ideal objective is intractable. 
Similar as the global parameters optimization, we can approximate the ideal objective as the empirical risk over training nodes. 
However, it can easily suffer from the over-fitting issue, which is shown on Figure \ref{overfit}.
We believe the empirical risk over validation nodes is a better approximation of the ideal objective since it reflects to what extent the parameter generalizes well.
In this light, we set $l_{outer}$ as the classification loss and optimize $\bm{Z}$ over the validation set $\mathcal{V}_{val}$ by minimizing Eq.~(\ref{outer}).
The details of updating $\bm{Z}$ is shown on line 8-11 in $\textbf{Algorithm}$ 1 and the mathematical formulation is shown on Appendix B.
With the parameter $\bm{Z}$ updated by the graph structure optimization, we reoptimize the $\bm{\mathcal{W}}$ with $\bm{Z}$ by global parameters optimization, and repeat this process until the early stopping is met.
To summarize, \textbf{Algorithm} \ref{alg} shows the training procedure of GSEBO. Besides, we explain why the bi-level optimization could defeat the over fitting problem when training on train and validation dataset on Appendix C.


\section{Experiments}
\begin{table}
\small
    \caption{Performance comparison across GNN architectures.}
    \label{result1}
     \vspace{-0.3cm}
    \centering
    \setlength\tabcolsep{5pt}
    \begin{tabular}{l|cccc}
      \toprule
      \multicolumn{1}{l|}{Method}&\multicolumn{1}{c}{Cora}&\multicolumn{1}{c}{Citeseer}&\multicolumn{1}{c}{Terrorist}&\multicolumn{1}{c}{Air-USA} \\
    \specialrule{0.1em}{0pt}{0pt}
    \rowcolor{lightgray} \multicolumn{5}{c}{GCN}\\
      
        Vanilla    &81.6 $\pm$ 0.7   &71.6 $\pm$ 0.4   &70.0 $\pm$ 1.1   &56.0 $\pm$ 0.8  \\
        AdaEdge     &81.9 $\pm$ 0.7   &72.8 $\pm$ 0.7   &71.0 $\pm$ 1.9   &57.2 $\pm$ 0.8 \\
        DropEdge    &82.0 $\pm$ 0.8   &71.8 $\pm$ 0.2   &70.3 $\pm$ 0.9   &56.9 $\pm$ 0.6\\
        GAUG        &83.2 $\pm$ 0.7   &73.0 $\pm$ 0.8   &71.4 $\pm$ 2.0   &57.9 $\pm$ 0.4 \\
        $\textbf{GSEBO}$   &$\textbf{84.0 $\pm$ 0.4}$   &$\textbf{74.4 $\pm$ 0.5}$   &$\textbf{72.1 $\pm$ 0.6}$   &$\textbf{59.8 $\pm$ 0.6}$    \\
      \specialrule{0.1em}{0pt}{0pt}
      \rowcolor{lightgray} \multicolumn{5}{c}{GAT}\\
        Vanilla    &81.3 $\pm$ 1.1   &70.5 $\pm$ 0.7   &67.3 $\pm$ 0.7   &52.0 $\pm$ 1.3\\
        AdaEdge     &82.0 $\pm$ 0.6   &71.1 $\pm$ 0.8   &$\textbf{72.2 $\pm$ 1.4}$   &54.5 $\pm$ 1.9     \\
        DropEdge    &81.9 $\pm$ 0.6   &71.0 $\pm$ 0.5   &69.9 $\pm$ 1.1   &52.8 $\pm$ 1.7    \\
        GAUG        &81.6 $\pm$ 0.8   &69.9 $\pm$ 1.4   &68.8 $\pm$ 1.1   &53.0 $\pm$ 2.0      \\
        $\textbf{GSEBO}$    &$\textbf{82.9$\pm${0.1}}$  &$\textbf{72.1$\pm$0.8}$  &69.7$\pm$1.6  &$\textbf{57.0$\pm$0.3}$    \\
      \specialrule{0.1em}{0pt}{0pt}
      \rowcolor{lightgray} \multicolumn{5}{c}{GraphSAGE}\\
      Vanilla    &81.3 $\pm$ 0.5   &70.6 $\pm$ 0.5   &69.3 $\pm$ 1.0   &57.0 $\pm$ 0.7     \\
        AdaEdge     &81.5 $\pm$ 0.6   &71.3 $\pm$ 0.8   &72.0 $\pm$ 1.8   &57.1 $\pm$ 0.5    \\
        DropEdge    &81.6 $\pm$ 0.5   &70.8 $\pm$ 0.5   &70.1 $\pm$ 0.8   &57.1 $\pm$ 0.5     \\
        GAUG        &$\textbf{81.7 $\pm$ 0.3}$   &71.4 $\pm$ 1.0   &70.4 $\pm$ 0.5   &55.0 $\pm$ 1.1     \\
        $\textbf{GSEBO}$    &80.4$\pm$0.9  &$\textbf{73.0$\pm$0.5}$  &$\textbf{76.5$\pm$0.9}$  &$\textbf{59.0$\pm$1.3}$  \\
      \specialrule{0.1em}{0pt}{0pt}
      \rowcolor{lightgray} \multicolumn{5}{c}{JK-Net}\\
      Vanilla    &78.8 $\pm$ 1.5   &67.6 $\pm$ 1.8   &70.7 $\pm$ 0.7   &53.1 $\pm$ 0.8   \\
      AdaEdge     &80.4 $\pm$ 1.4   &68.9 $\pm$ 1.2   &71.2 $\pm$ 0.7   &59.4 $\pm$ 1.0     \\
        DropEdge    &80.4 $\pm$ 0.7   &69.4 $\pm$ 1.1   &70.2 $\pm$ 1.3   &58.9 $\pm$ 1.4   \\
        GAUG        &79.4 $\pm$ 1.3   &68.9 $\pm$ 1.3   &70.2 $\pm$ 0.5   &52.3 $\pm$ 1.8    \\
        $\textbf{GSEBO}$    &$\textbf{81.8$\pm$1.0}$  &$\textbf{69.4$\pm$1.0}$  &$\textbf{73.7$\pm$1.2}$  &$\textbf{59.6$\pm$1.2}$ \\
      
                        
      \bottomrule
    \end{tabular}
    \vspace{-0.2cm}
  \end{table}
  
In this section, we conduct experiments on four datasets to answer the following research questions: (1) how does the performance of the proposed GSEBO compared with the state-of-the-art methods? (2) how robust is the proposed GSEBO? (3) to what extent does the proposed GSEBO decrease the impact of inter-class connections? (4) what are the factors that influence the effectiveness of the proposed GSEBO? We present the results of questions (3) and (4) on Appendix D and E due to space limitations.

  
    

\subsection{Experimental Setup}
\paragraph{Dataset.} We select four widely used node real-world classification benchmark datasets with graph of citation networks (Cora and Citeseer~\cite{kipf2016semi}), 
social networks (Terrorist)~\cite{Zhao2006EntityAR}, and air traffic (Air-USA)~\cite{WuHX19}. 
We summarize the statistics of the datasets and describe in detail on Appendix F.  
We adopt the same data split of Cora and Citeseer as \cite{kipf2016semi}, and a split of training, validation, testing with a ratio of 10:20:70 on other datasets~\cite{zhao2020data}. 

\paragraph{Compared methods.} We apply GSEBO on 4 representative GNN architectures: GCN \cite{kipf2016semi}, GAT \cite{velivckovic2017graph},
GraphSAGE \cite{hamilton2017inductive} and JK-Net \cite{xu2018representation}. For each GNN, we compare GSEBO with 
the vanilla version, and three variants with state-of-the-art connection modeling methods: AdaEdge \cite{chen2020measuring}, 
DropEdge \cite{rong2019dropedge} and GAUG
\cite{zhao2020data}.
In addition, 
we compared the GSEBO with GSL methods: BGCN
\cite{zhang2019bayesian}, 
VGCN
\cite{elinas2019variational}, 
PTDNet
\cite{luo2021learning}, and advanced attention mechanism: MAGNA
\cite{wang2021multihop}.
Note that the base model of GSEBO, BGCN, and PTDNet are GCN.

\paragraph{Implementation details.} 
In the experiments, the latent dimension of all the methods is set to the 16. The parameters for all baseline methods are initialized as the corresponding papers, and are carefully tuned to achieve optimal performances. The learning rate of inner and outer optimization are set to 0.01. The hyperparameter $\lambda$ is set to $5\times{10^{-4}}$, and we search for the inner learning depth $\tau$ over the range [5,10,15,20,25]. Due to bilevel optimization can not guarantee the convergence, we set the patience of early stopping to 20 to terminate the optimization of GSEBO. To prevent overfitting, the dropout ratio is set to 0.5 for all the methods. 
The network architectures of all the mehtods are configured to be the same as described in the original papers.
Our experiments are conducted with Tensorflow running on GPU machines (NVIDIA 2080Ti). For all the compared methods, we report the average accuracy on the test set over 10 runs.

\begin{table}
\small
    \caption{Performance comparison of GSEBO with GSL methods.}
    \label{result2}
    \vspace{-0.3cm}
    \centering
    \setlength\tabcolsep{5pt}
    \begin{tabular}{l|cccc}
      \toprule
      \multicolumn{1}{l|}{Method}&\multicolumn{1}{c}{Cora}&\multicolumn{1}{c}{Citeseer}&\multicolumn{1}{c}{Terrorist}&\multicolumn{1}{c}{Air-USA} \\
    
      \specialrule{0.1em}{0pt}{1pt}
      \multicolumn{1}{l|}{BGCN}      &81.2 $\pm$ 0.8     &72.4 $\pm$ 0.5      &70.3$\pm$0.8      &56.5 $\pm$ 0.9 \\
      \multicolumn{1}{l|}{VGCN}      &64.4 $\pm$ 0.2     &67.8 $\pm$ 0.8      &$\textbf{73.8 $\pm$ 0.9}$    &53.3 $\pm$ 0.3      \\
      \multicolumn{1}{l|}{PTDNet}      &82.8 $\pm$ 2.6     &72.7 $\pm$ 1.8      &68.3 $\pm$ 1.6    &53.4 $\pm$ 1.4      \\
      \multicolumn{1}{l|}{MAGNA}      &81.7 $\pm$ 0.4     &66.4 $\pm$ 0.1      &67.2$\pm$0.1      &55.1 $\pm$ 1.2     \\
      \multicolumn{1}{l|}{$\textbf{GSEBO}$}      &$\textbf{84.0$\pm$0.4}$    &$\textbf{74.4$\pm$0.5}$     &72.1$\pm$0.6   &$\textbf{59.8$\pm$0.6}$       \\
                        
      \bottomrule
    \end{tabular}
  \end{table}
  
\subsection{Performance Comparison}
Table \ref{result1} shows the performance of GSEBO instantiate with classical GNNs, and Tabel \ref{result2} presents the results comparison of GSEBO and state-of-the-art GSL methods.
From Table \ref{result1} and Table \ref{result2}, we have the following observations: 

\begin{figure}
  \centering
  \setlength{\abovecaptionskip}{0cm}
  \setlength{\belowcaptionskip}{0cm}
  \includegraphics[scale=0.48]{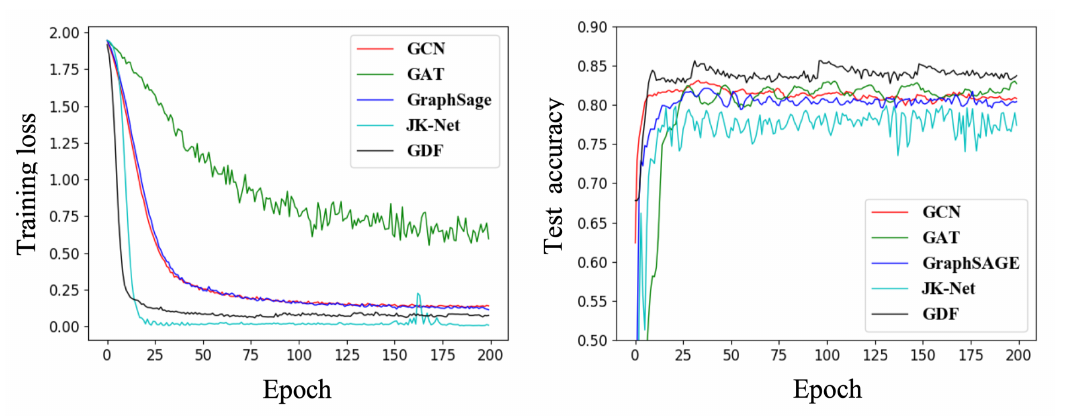}
  \caption{The training loss and test accuracy of classical GNN and GSEBO on Cora.}
  \vspace{-0.4cm}
  \label{loss}
\end{figure}

\begin{figure*}
  \centering
  \setlength{\abovecaptionskip}{0cm}
  \setlength{\belowcaptionskip}{0cm}
  \includegraphics[scale=0.62]{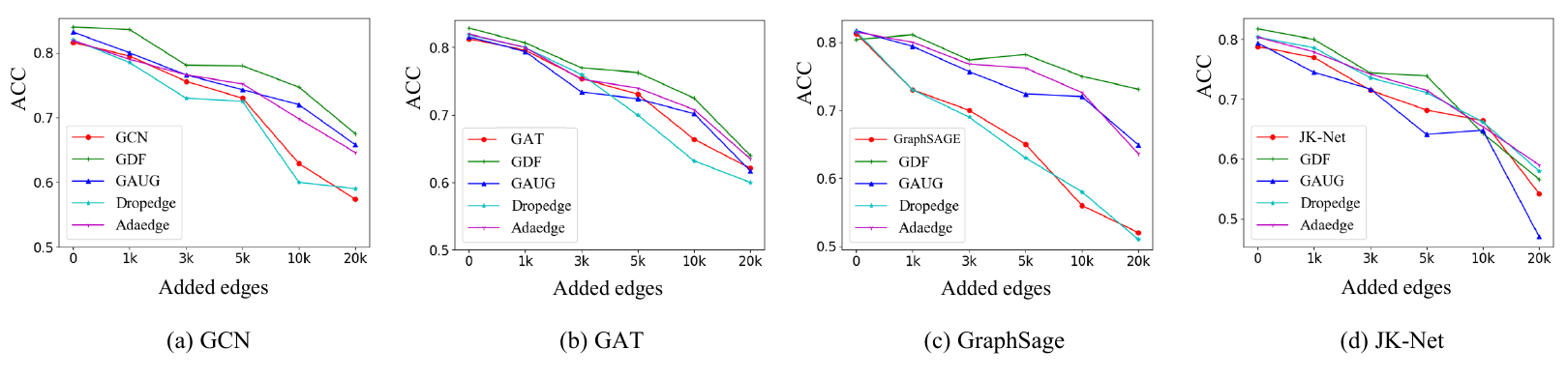}
  \caption{Node classification performance of GSEBO on Cora poisoned at different inter-class levels.}
  \vspace{-0.4cm}
  \label{noisyedge}
\end{figure*}

\paragraph{Improvement over baselines.}
GSEBO outperforms the baselines in most cases. Considering the average performance over four datasets, the improvement of GSEBO over the baselines is in the range of [3.0\%, 12.5\%], which validates the effectiveness of the proposed method.
In particular, (1) \textbf{Probabilistic mechanisms}. The performance gain of BGCN and VGCN over the vanilla GCN are limited, which might because of the unsatisfied assumption of the prior distribution. This result shows the rationality of relaxing the prior assumption and modeling the connection strength with parameters directly. 
(2) \textbf{Connection modeling}. DropEdge, AdaEdge, GAUG, and PTDNet achieve better performance than vanilla GNN, which modify the structure of the graph from different perspectives such the smoothness~\cite{chen2020measuring} and robustness~\cite{rong2019dropedge}. These results reflect the benefit of connection modeling. However, there is still a clear gap between these methods and GSEBO, which is attributed to the global objective for learning the edge strength.
(3) \textbf{Attention mechanism}. The performance of MAGNA is inferior, which is consistent with the analyzing in~\cite{knyazev2019understanding}.

\paragraph{Effects across GNN architectures.} 
On the four GNN architectures, GSEBO achieves better performance than the vanilla version in all cases. In particular, GSEBO achieves an average improvement across datasets of 4.2\%, 4.4\%, 4.0\%, and 5.74\% over the vanilla GCN, GAT, GraphSAGE, and JK-Net, respectively. These results justify the effectiveness of structure learning of the proposed GSEBO. 
Across the four architectures, GSEBO achieves the largest improvement over JK-Net. We postulate the reason is that the jump connection in JK-Net makes it aggregates more hops of neighbors than the other GNNs. As the hops increase, the homophily ratio of neighbors will decrease, \ie more neighbors are in classes different from the target node. Therefore, optimizing the connection strength (\ie $\bm{Z}$) is more beneficial on JK-Net.

\paragraph{Effects across datasets.}
From the perspective of dataset, GSEBO consistently performs better than the vanilla version on the four datasets. Specifically, 
 the average improvement over the four across classic GNN architectures achieved by GSEBO are 1.9\%, 3.1\%, 5.3\% and 8.0\% on the four datasets, respectively.
Considering that the four datasets come from different scenarios, these results are evidence for the potential of GSEBO to be widely applied in practice.
Note that the trend of performance improvement is similar to the density of graph where Air-USA is the most dense graph with the largest performance improvement. 
As the number of neighbors increases, the percentage of neighbors essential for the classification of the target node will decrease. 
This result can reflect the rationality of optimizing the connection strength according to the overall classification objective.

Moreover, the training loss and test accuracy of GNN methods and the proposed GSEBO are shown in Figure \ref{loss}. We observe that the loss of GSEBO tend to be stable after 30 epochs and is smaller than GCN, GAT, and GraphSAGE, showing the empirical convergence of GSEBO. This is because simultaneously optimizing the parameters of $\bm{\mathcal{W}}$ and $\bm{Z}$ would make the loss function more easily approach to the labels. However, the loss of JK-Net is smaller than GSEBO, we attribute it to that JK-Net is easier to trap in over-fitting.

\subsection{Robustness Analysis}
We investigate the robustness of GSEBO under the different inter-class levels. In particular, we follow \cite{luo2021learning} and construct graphs based on Cora by randomly adding 1,000, 3,000, 5,000, 10,000, and 20,000 inter-class edges, respectively.
On the synthetic datasets, we compare GSEBO with Vanilla, GAUG, DropEdge, and AdaEdge.
Figure \ref{noisyedge} shows the performance on the five GNN architectures. From the figures, we have the following observations:
\begin{itemize}[leftmargin=*]
    \item The margins between GSEBO and vanilla GNN on the synthetic datasets are larger than the original Cora. For example, when adding 20,000 inter-class edges, GSEBO improves the accuracy by 17.6\%, 3.1\%, 40.6\%, and 4.4\% over GCN, GAT, GraphSAGE and JKNet. This result indicates the robustness of GSEBO's structure learning ability.

    \item In most cases, GSEBO outperforms the baselines at different noisy levels, which further justifies its  robustness.
    
    \item GAUG and AdaEdge utilizes different strategies to update the structure of the graph, which also consistently perform better than vanilla GNN. However, their gaps to GSEBO on the synthetic data are larger than the original Cora. We postulate that the reason is their objectives are affected by the intentionally added noise.
    \item DropEdge shows worse performances than the vanilla GNN on the synthetic datasets. The comparison shows that randomly dropping edges fails to enhance GNN robustness when the noisy level is high.
\end{itemize}

\section{Conclusion}
In this work, we propose a novel GSEBO, which utilizes the global information to learn the graph structure. To better optimizes the strength of each edge and the parameters of feature mapping, we decompose the optimization problem into two mutually constrained optimization objectives, \ie the inner and outer objective based on a a universal graph convolution operator. 
Extensive experiments demonstrate the effectiveness and robustness of GSEBO on both benchmark and synthetic datasets.

Although the impressive results GSEBO achieved, there are still some limitations: GSEBO cannot be effectively applied to large graphs, which requires implementation of mini-batches. Besides, we evaluate GSEBO in the transductive setting, when new nodes add to the graph after training, GSEBO has to retrain the entire model.
For future researches, we would like to explore solutions to the above limitations.
\clearpage
\balance
\bibliographystyle{named}
\bibliography{ijcai22}

\end{document}